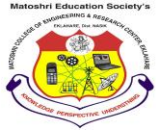
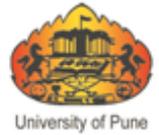

# Clustering Multidimensional Data with PSO based Algorithm

Jayshree Ghorpade-Aher and Vishakha A. Metre

**Abstract**: Data clustering is a recognized data analysis method in data mining whereas K-Means is the well known partitional clustering method, possessing pleasant features. We observed that, K-Means and other partitional clustering techniques suffer from several limitations such as initial cluster centre selection, preknowledge of number of clusters, dead unit problem, multiple cluster membership and premature convergence to local optima. Several optimization methods are proposed in the literature in order to solve clustering limitations, but Swarm Intelligence (SI) has achieved its remarkable position in the concerned area. Particle Swarm Optimization (PSO) is the most popular SI technique and one of the favorite areas of researchers. In this paper, we present a brief overview of PSO and applicability of its variants to solve clustering challenges. Also, we propose an advanced PSO algorithm named as Subtractive Clustering based Boundary Restricted Adaptive Particle Swarm Optimization (SC-BR-APSO) algorithm for clustering multidimensional data. For comparison purpose, we have studied and analyzed various algorithms such as K-Means, PSO, K-Means-PSO, Hybrid Subtractive + PSO, BRAPSO, and proposed algorithm on nine different datasets. The motivation behind proposing SC-BR-APSO algorithm is to deal with multidimensional data clustering, with minimum error rate and maximum convergence rate.

## I. INTRODUCTION

SEARCHING for useful nuggets of information amongst huge amount of data is known as the field of *data mining*. Data mining system is the tool for extracting any hidden but meaningful information in data that facilitates its use for decision-making processes. On the other side, Artificial Intelligence has proved itself as a great tool in recent years, with the help of which computer can make decisions without considering any input from human being. AI inspired techniques impart a 'sixth sense' to the data mining systems and explore meaningful information. The combination of these two astonishing domains, have motivated us to utilize their outstanding features and apply them to optimize any complex real world problems.

Swarm intelligence is a relatively new subfield of AI, which studies an emergent collective intelligence of groups of simple agents [23]. It is based on social behavior that can be observed in nature, such as ant colonies, bee hives, fish schools, and flocks of birds, where a number of individuals with limited capabilities are able to come to intelligent solutions for complex problems [19]. In recent years Particle Swarm Optimization (PSO) which is one of the swarm intelligence paradigms that have received widespread attention in research [24]. PSO [10] is an evolutionary computation technique which simulates the movement of flock of birds that performs a global search within a solution space [16],[17],[18]. PSO produces better results in complicated and multipeak problems with few parameters to adjust giving fast as well as accurate computation results which lead to be a popular optimization technique in swarm intelligence field [13],[18].

Data mining is one of the vast fields and has variety of application areas to adapt as our research work. But, data clustering which is the well known data analysis method in data mining, achieved its remarkable position among all [20]. It is the method whose basic aim is to group the data objects into meaningful groups, possessing *minimum intra cluster distance* and *maximum inter cluster distance* [6],[7]. Clustering has its own importance in variety of applications nowadays such as biomedical and biological sequences, collective behavior, image segmentation, power engineering, WSN, etc. But at the same time it presents several challenges such as anomaly detection, dimensionality reduction, feature extraction, grouping, similarity computation, etc. making it as a favorite area of researchers.

The rest of the paper is organized as follows. In Section II, a brief literature survey on PSO and its different variants is described. Section III provides implementation details of our project. Section IV analyses the different datasets and expected results, followed by Section V which concludes this paper providing future direction.

## II. LITERATURE SURVEY

It is an undeniable fact that all of us are optimizers and make decisions for the sole purpose of maximizing our quality of life, productivity in time, as well as our welfare in some way or another. Optimization is defined as a methodology to determine the best-suited solution for any problem under given state of affairs [26]. Optimization plays very important role in almost every area of applications but data clustering has become a promising area of research work these days. Different researchers are proposing different approaches for optimizing clustering challenges. Ant Colony Optimization (ACO) [23],[25], Genetic Algorithm (GA) [11],[25],[27], Particle Swarm Optimization (PSO) [18],[19],[23],[25], and Stimulated Annealing (SA) [25] are most popular optimization techniques till date. The basic classification of optimization techniques is studied and analyzed as depicted in the Fig. 1.

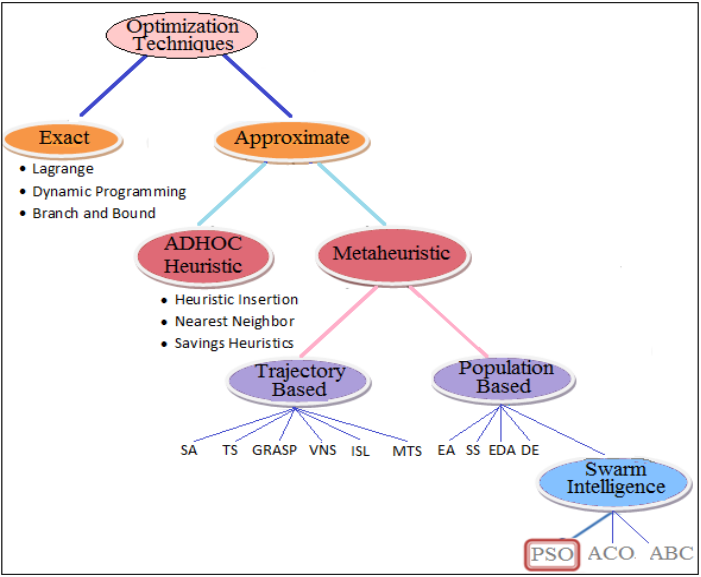

Fig. 1. Analyzed Classification of Optimization Techniques

Table I gives a brief comparative analysis of four promising optimization techniques, while the survey on applicability of different PSO variants to data clustering is described in Table II.

### III. Implementation Details

Data clustering with PSO algorithms have recently been shown to produce good results in a wide variety of real-world data. Although the variety of variations in PSO based clustering techniques are proposed in literature achieving better results, suffers with several limitations more specifically when dealing with multidimensional data [2],[4],[22]. This research presents a novel PSO based clustering technique which is a hybrid approach to the multidimensional data clustering problem by predicting the initial cluster center location and number of clusters and optimizing it with one of efficient PSO variant. A hybridized approach involves swarm intelligence inspired algorithm (i.e. BRAPSO) [1] which is a new variant of PSO with Subtractive clustering algorithm [2],[21] that evaluates clustering over multidimensional data.

#### A. Architecture/Block Diagram

In proposed approach, Subtractive clustering is followed by BRAPSO which is sequentially termed as Subtractive Clustering based Boundary Restricted Adaptive PSO (SC-BR-APSO) algorithm. This is basically an improvement over hybrid Subtractive clustering + Simple PSO algorithm [2]. Thus, SC-BR-APSO architecture includes two modules, i.e. Subtractive clustering module and BRAPSO module as shown in Fig. 2.

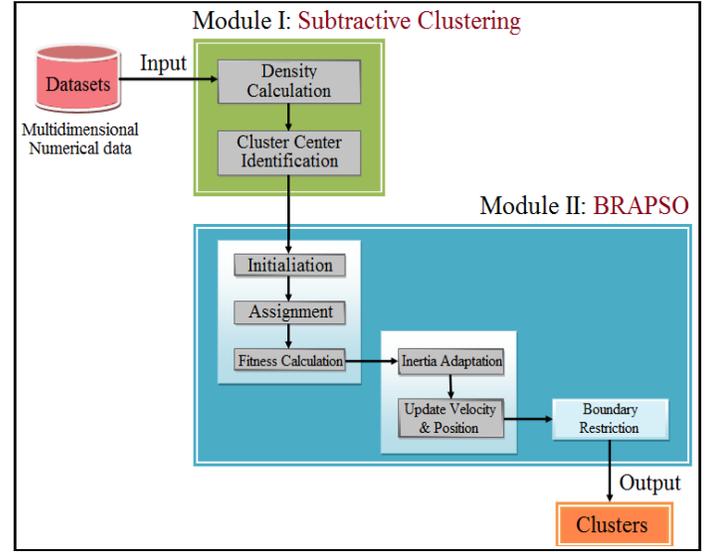

Fig. 2. Architecture of Proposed System

*1) Module I: Subtractive Clustering*

In subtractive clustering, a problem space is the multidimensional vector space. Each dot represents a dimension vector in the problem space. Hence the entire dataset represents multidimensional space with a large number of dots in the search space [2]. Such multidimensional data points will be served as input to the algorithm.

At the initial stage, the Subtractive clustering module is executed to calculate density function value for each data point in the search space. The computational complexity is linearly proportional to the number of data points in the space and independent of the dimensions of the considered problem [2].

Consider a multidimensional space having collection of *n* data points *{x₁,…, xₙ}*. Since each data point is a candidate for cluster centers, a density measure at data point *xᵢ* is calculated by using (1) [2],

$$D_i = \sum_{j=1}^{n} exp\left(-\frac{\|x_i - x_j\|^2}{(r_a/2)^2}\right) \quad (1)$$

*Where,*
$D_i$ - *density value of $i^{th}$ data point*
$x_i$ - $i^{th}$ *data point*
$x_j$ - $j^{th}$ *data point*
$r_a$ - *neighborhood radius (positive constant)*

Hence, once we are done with the calculation of the density measure for all data points, the one with the highest density measure is elected as the first cluster center. To derive other cluster centers, the density measure $D_i$, for each data point $x_i$ are recalculated by using (2) [2],

$$D_i = D_i - D_{c_1} exp\left(-\frac{\|x_i - x_{c_1}\|^2}{(r_b/2)^2}\right) \quad (2)$$

*Where,*
$D_{c_1}$ - *density value of first cluster center*
$x_{c_1}$ - *first cluster center*
$r_b$ - *neighborhood radius; having measurable reduction in the density measure [2] (positive constant)*

Track 8: Soft Computing and Artificial Intelligence

TABLE I
COMPARATIVE ANALYSIS OF OPTIMIZATION TECHNIQUES

| Technique | Type | Basic Concept | PROS | CONS |
|---|---|---|---|---|
| ACO | Swarm Intelligence Method | It used to discover the shortest trail to the source of food from the colony and return back by means of an indirect communication through pheromone. | 1. Better performance (when compared to GA and SA). 2. Retains memory of entire colony instead of previous generation only. 3. Less affected by poor initial solutions. 4. Robust and also easy to accommodate with other algorithms. | 1. For large number of nodes, computationally very difficult to solve. 2. Convergence is guaranteed, but time to convergence is uncertain. 3. Complicated coding. 4. Tradeoffs in evaluating convergence. |
| GA | Evolutionary Algorithms | Endurance of the best approach for selecting the best solution from the available solutions. | 1. Supports multi-objective optimization. 2. Effective for large, complex and poorly known search space. 3. Protection against early convergence to local optimum. | 1. Difficult to operate on dynamic data. 2. Tend to converge towards local optima and needs complex operators. 3. Inappropriate to optimize constraint based problems. |
| PSO | Swarm Intelligence Method | Optimization of non linear functions using methodology of particle swarms (i.e. bird flocking behavior). | 1. Simple, easy and derivative free algorithm. 2. Very few parameters to adjust and efficient global search ability. 3. Efficient to handle complex non linear optimization problems. 4. They have internal memory and each particle represents individual solution. 5. Best information sharing mechanism. | 1. For large search space, premature convergence to local optima. 2. Weak local search. |
| SA | Probabilistic meta-heuristic algorithm | It is the simulation of evolution of a solid in a heat bath to thermal equilibrium. | 1. Guarantees finding an optimal solution. 2. Easy to code for complex problems. 3. It is versatile as it does not depend on any restrictive properties of the model. | 1. Very costly and time consuming. 2. Not much useful for search space with few local minima's. |

TABLE II
COMPARATIVE ANALYSIS OF DIFFERENT PSO VARIANTS FOR DATA CLUSTERING

| Publication | Paper Title | Purpose | Algorithm/ Techniques | Datasets | PROS | CONS |
|---|---|---|---|---|---|---|
| Springer 2012 S. Rana et al.[1] | A boundary restricted adaptive particle swarm optimization for data clustering | To apply Boundary restriction strategy for improved clustering. | BRAPSO | Art1, Art2, CMC, Crude oil, Glass, Iris, Vowel, and Wine. | 1. Efficient, robust and fast convergence. 2. Improved clustering by handling particles outside search space boundary. | - |
| IJORCS 2013 Mariam El-Tarabily et al.[2] | A PSO-Based Subtractive Data Clustering Algorithm | To enhance clustering speed. | Hybrid Subtractive clustering PSO | Iris, Wine, Yeast | 1. High convergence speed and minimum fitness value. 2. No need of any dimension reduction approach. | 1. Overlapping cluster membership problem. 2. Premature convergence problem. |
| IJSER June 2013 Chetna Sethi et al.[3] | A Linear PCA based hybrid k-Means PSO algorithm for clustering large dataset | To cluster high dimensional data. | PCA-K-PSO, k-means, Linear PCA, PSO | Artificial data sets | 1. Compact clustering. 2. Reduced computational cost. | 1. Dimension reduction may degrade clustering results. |
| IEEE Transaction April 2010 Serkan Kiranyaz et al.[4] | Fractional Particle Swarm Optimization in Multidimensional Search Space | To address premature convergence problem for multidimensional data. | MD-PSO with FGBF, Fuzzy clustering, FGBF, MD-PSO | Artificial data sets | 1. Suitable for multi dimensional datasets. 2. Avoids premature convergence to local optima. | 1. Occasional over-clustering can be encountered. 2. Increased dimension increases cost. |
| Springer July 2010 Bara'a Ali Attea [5] | A fuzzy multi-objective particle swarm optimization for effective data clustering | To address the problem of multicluster membership. | FMOPSO, Multiobjective optimization Technique, Fuzzy clustering | Cancer, Iris, Soybean, Tic-tac-toe, Zoo | 1. Reliable and efficient clustering algorithm. 2. Good clustering result. | 1. Exhibits problem with multidimensional datasets. |

Generally, $r_b$ is equal to *1.5 $r_a$*, as suggested in [2]. When the density measure for each data point is recalculated, the next cluster center $x_{c_2}$ is selected and the density measures for all data points are again recalculated. This process is iteratively repeated until a sufficient number of cluster centers are obtained. These cluster centers would be convincingly used as the initial cluster centers for the next module.

*2) Module II: Boundary Restricted Adaptive PSO (BRAPSO)*

The output from subtractive clustering module i.e. initial cluster center and number of clusters will be given to BRAPSO module as an initial seed. *Module II* starts with basic PSO process, in which the birds in a flock are symbolically represented by particles. These particles are simple agents "flying" through a problem space, represents individual solution to problem [14],[15],[18],[23]. A different problem solution is generated, as a particle updates itself to new location. The fitness function is evaluated for each particle in the swarm and compared with the fitness of its own best previous position i.e. $p_{best}$ and to the fitness of the global best particle amongst all particles in the swarm i.e. $g_{best}$. Then, the velocities and positions for the $i^{th}$ particle are updated by using (3) & (4) [2],[18],[23],

$$v_{id} = w * v_{id} + c_1 * rand_1 * (p_{best} - x_{id}) + c_2 * rand_2 * (g_{best} - x_{id}) \quad (3)$$

$$x_{id} = x_{id} + v_{id} \quad (4)$$

Where,
d - dimension of the problem space
$rand_1$, $rand_2$ - random values in the range of (0, 1)
$c_1$, $c_2$ - acceleration coefficients constants
w - inertia weight factor

The necessary diversity of the swarm is provided by the inertia weight factor *w*, by changing the momentum of particles to avoid the stagnation of particles at the local optima [2]. The previous work adopted linear decrease in the inertia weight but we observed better solutions or swarm crowding near the global solution with small change in the inertia weight and hence motivated us to adapt non-linear (exponential) inertia weight [1]. The exponential decrease in inertia explores the entire search area and guarantees the global convergence. The value of the inertia weight can be calculated by using (5) [1],

$$w = maxw * (exp(-iter)) \quad (5)$$

Where,
maxw - maximum value of w (i.e. 0.9)
iter - current iteration number

When evaluation of the clustering process is started, some particles have tendancy to travel beyond the search space area, which may cause inaccurate clutsering. This problem of PSO can be handled by boundary restriction strategy [1] as follows:

/* boundary restrcition strategy*/
If position <= upbnd and position >= lwbnd
   Then position = position
Else position = position - velocity;

Finally, the points are assigned to their closest cluster and cluster centers are recalculated. This process is repeated untill the convergence occurs, giving the optimal clusters as a clustering solution.

*B. Proposed SC-BR-APSO Algorithm*

In the Subtractive Clustering based Boundary Restricted Adaptive Particle Swarm Optimization Algorithm, the Subtractive clustering algorithm is used at the initial stage to help discovering the neighborhood of the optimal solution by suggesting good initial cluster centers and the number of clusters. This output of Subtractive algorithm is used as the initial seed to the BRAPSO algorithm, which is applied for refining and generating the final result using the global search capability of BRAPSO. The pseudo code of proposed SC-BR-APSO algorithm is given in Fig. 3.

---

**Algorithm of the proposed scheme SC-BR-APSO**

1. Consider any *M*-dimensional data set having *N* particles.
2. Calculate density function

$$D_i = \sum_{j=1}^{n} exp\left(-\frac{\|x_i - x_j\|^2}{(r_a/2)^2}\right), i = 1, \ldots, n, j = 1, \ldots, z$$

/* Where *n* and *z* are numbers of data sets and clusters, respectively and $x_i$ and $x_j$ are $i^{th}$ and $j^{th}$ data points, respectively and $D_i$ is the density value of $i^{th}$ data point and $r_a$ is the neighborhood radius (positive constant). */

3. If sufficient numbers of clusters then identify initial cluster centers and proceed else go to step 2.
4. Randomly initialize particle velocity and position.
5. Calculate fitness function and set $p_{best}$ and $g_{best}$.
6. Calculate the value of inertia weight factor *w* exponentially
$w = maxw * (exp(-iter))$
/* Where maximum value of *w* (i.e. 0.9) and iter is current iteration number. */
7. Update velocity and position of particle with boundary restriction strategy
  *If position <= upbound and position >= lwbound*
    *Then position = position*
  *Else position = position - velocity;*
/* Where *position* and *velocity* are the current position and velocity of particle, respectively and *upbound* and *lwbound* are the upper limit and lower limit specified for search space area. */
8. Assign particles to their nearest cluster center and recalculate the cluster centers.
9. Repeat steps until convergence.

---

Fig. 3. Proposed SC-BR-APSO algorithm

*C. Platform*

Platform: Windows (XP/Vista/7/8 etc)
Tools for Programming: Microsoft Visual Studio.NET 2010
Hardware: PC/Laptop, 64 bit processors (Core$^{TM}$ i3 processor)
       Ram 256 MB, Hard disc space 512 MB.
Technology: C#.Net

## IV. RESULT ANALYSIS

*A. Experimental Setup*

The proposed research work is designated to perform comparative analysis on 10 well known datasets in order to test efficiency of K-Means algorithm, PSO, K-Means-PSO, Hybrid Subtractive + PSO, BRAPSO and proposed SC-BR-APSO algorithm. The 9 datasets, which we are intended to use are selected from the real datasets of UCI and are easily available at link *http://archive.ics.uci.edu/ml/datasets.html*.

These datasets are namely Breast Cancer Wisconsin, Contraceptive Method Choice (CMC), Crude Oil, Glass, Iris, Pima Indians Diabetes, Vowels, Wine, and Zoo. Each & every considered data sets are having different number of dimensions/attributes (e.g. high, low, and medium), which promotes them as important multidimensional datasets for experimental studies [9],[10]. The detailed characteristics of these 9 data sets [1],[2],[9],[10] are well illustrated in Table III as given below:

TABLE III
CHARACTERISTICS OF DATA SETS

| Dataset | Dimension Type | Total no. of instances | No. of Clusters | No. of instances in each class | No. of Dimensions |
|---|---|---|---|---|---|
| Cancer | Integer | 683 | 2 | 444,239 | 9 |
| CMC | Integer | 1473 | 3 | 629,334, 510 | 9 |
| Crude Oil | Integer | 56 | 3 | 7,11,38 | 5 |
| Glass | Real | 214 | 6 | 70,17,76, 13,9,29 | 9 |
| Iris | Real | 150 | 3 | 50,50,50 | 4 |
| Pima | Integer, Real | 768 | 2 | 500, 268 | 8 |
| Vowel | Numerical | 871 | 6 | 72,89, 172,151, 207,180 | 3 |
| Wine | Integer, Real | 178 | 3 | 59, 71, 48 | 13 |
| Zoo | Integer | 101 | 7 | 41, 20, 5, 13, 4, 8, 10 | 17 |

To measure the performance of the different considered algorithms, we are using three criteria's namely *Sum of Intra Cluster Distances (SICD)*, *Error rate*, and *Convergence rate*. The distance measure of each data point within a cluster to its cluster center is termed as *SICD* and minimum SICD leads to higher quality clustering result. Hence in case of data clustering, the basic aim of any PSO variant is to optimize the fitness function as minimum as possible [9]. The objective function in (6) will act as a fitness function in PSO algorithm.

$$J(C_1, C_2, \ldots, C_K) = \sum_{i=1}^{K} \left( \sum_{X_j \in C_i} \|Z_i - X_j\| \right) \quad (6)$$

*Where,*
*K - number of clusters*
*N - data points/vectors*
$Z_i$ - $i^{th}$ *cluster center* $(1 \leq i \leq K)$
$X_j$ - $j^{th}$ *data point* $(1 \leq j \leq N)$

The total number of misplaced points over the entire dataset points is known as an error in the clustering that can be calculated by using (7) [9],

$$Error = \frac{1}{N}\left(\sum_{i=1}^{N}(if\ (Class(i) = Cluster(i))\ then\ 0\ else\ 1)\right) \times 100 \quad (7)$$

*Where,*
*N - total number of the points in dataset.*
*Class(i) - ith class*
*Cluster(i) - ith cluster*

For fast global convergence, we must propose such a variant of PSO that will achieve high convergence rate. Hence, we are adapting non-linear (exponential) inertia weight factor [9].

### B. Expected Results

Experimental testing for results of proposed SC-BR-APSO algorithm on 9 different datasets is performed & analyzed according to following steps:

1. Load any multidimensional numerical dataset (e.g. the clustering problem with the data points shown in Fig. 4. (a).)

2. Start subtractive clustering algorithm to find initial cluster centers and number of clusters as shown in Fig. 4. (b). (three clusters with $c_1$, $c_2$, and $c_3$ as the initial cluster centers). Assign particles to their nearest cluster centers with respect to any distance measure (i.e. Euclidean distance measure). While assigning the particles to the respective cluster center, some particles may share their membership with more than one cluster, which leads to inaccurate clusters. Hence, such particles are assigned to the appropriate cluster centers.

3. Begin APSO [12] process on the output given by the SC module. This will initiate the position and velocity of particles and the calculation of fitness function for each and every particle. While moving in the search space, $p_{best}$ and $g_{best}$ will get discovered and every particle updates their position as well as velocity according to them [8]. In order to find global optima, every particle follows gbest particle. Finally, apply boundary restricted strategy on the output of APSO to handle the particles travel beyond the search space while finding the global optima. This will put such particles back into the search space and recalculate cluster centers to obtain final optimal clustering solution as depicted in the Fig. 4. (c).

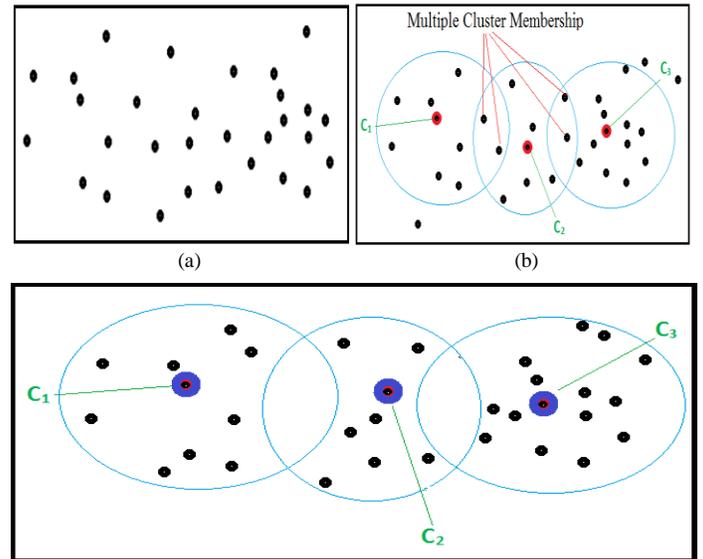

Fig. 4.(a) Initial data particles (b) Output of Subtractive Clustering algorithm (c) Final clustering solution (optimal).

## V. CONCLUSION AND FUTURE SCOPE

We have analyzed the different aspects of the applicability of several PSO variants to data clustering challenges. We have also proposed SC-BR-APSO algorithm, which is a hybrid of Subtractive Clustering and BRAPSO algorithms for clustering multidimensional data. Thus, the proposed approach is designed to overcome the major problems associated with the existing technique i.e. hybrid subtractive + PSO clustering algorithm as well as the other methodologies in the literature. Error rate, convergence rate, number of iterations, and Sum of Intra Cluster Distances (SICD) are four important aspects to be considered for comparison purpose. SC-BR-APSO will be tested on 9 different kinds of datasets having multiple dimensions and it is predicted to achieve compact clustering result (i.e. minimum SICD) with increased convergence performance and decreased error rate with minimum iterations as compared to existing algorithms. With the curse of dimensionality, we observed that time complexity is directly proportional to the size of the dataset; hence future work is intended to minimize the complexity of the algorithm while achieving the better performance.

**Jayshree Ghorpade-Aher** has completed her B.E. Computers and full-time M.E. Computers in 2004 from PICT, University of Pune. Her research areas include DM, IP, and Soft Computing and have published about 30 research papers. Awarded *"BEST PAPER AWARD"* by the ACM International Conference, CUBE' 12. Also, published 2 *National Patents* and wrote a *BOOK* on Computer Networks. She is currently working as a Subject Chairman, Pune University for ME Computers and Assistant Professor at MITCOE, Pune. She has guided nearly 50+ projects at UG and PG level.

**Vishakha A. Metre** has completed her B.E. Information Technology in 2010 from JDIET, SGB Amravati University. She is currently pursuing full-time M.E. Computers from MITCOE, University of Pune. She has published 2 International Journal Papers and her research areas include Artificial Intelligence, Data Mining, and Soft Computing. Recently she has won the *FIRST PRIZE* at the National Level Poster Presentation Event-Dexterity 2K14, organized by MMCOE, Pune.